\documentclass[letterpaper]{article} 
\usepackage{aaai2026}
\usepackage{times}  
\usepackage{helvet}  
\usepackage{courier}  
\usepackage[hyphens]{url}  
\usepackage{graphicx} 
\usepackage{natbib}  
\usepackage{caption} 
\usepackage{algorithm}
\usepackage{algorithmic}
\usepackage{amsmath}
\usepackage{amssymb}
\usepackage{booktabs} 
\usepackage{multirow}
\usepackage{graphicx}
\usepackage{subcaption}
\usepackage{float}
\usepackage{xcolor}
\usepackage{newfloat}
\usepackage{listings}
\DeclareCaptionStyle{ruled}{labelfont=normalfont,labelsep=colon,strut=off} 
\floatstyle{ruled}
\newfloat{listing}{tb}{lst}{}
\floatname{listing}{Listing}
\nocopyright 

\title{BSMamba: Brightness and Semantic Modeling for Long-Range Interaction in Low-Light Image Enhancement}
\author{
    Tongshun Zhang\textsuperscript{\rm 1,\rm 2},
    Pingping Liu\textsuperscript{\rm 1,\rm 2}\thanks{Corresponding author},
    Mengen Cai\textsuperscript{\rm 1,\rm 2},
    Zijian Zhang\textsuperscript{\rm 1,\rm 2},
    Yubing Lu\textsuperscript{\rm 1,\rm 2},
    Qiuzhan Zhou\textsuperscript{\rm 3}
}
\affiliations{
    \textsuperscript{\rm 1} College of Computer Science and Technology, Jilin University\\
    \textsuperscript{\rm 2} Key Laboratory of Symbolic Computation and Knowledge Engineering of Ministry of Education, Jilin University\\
    \textsuperscript{\rm 3} College of Communication Engineering, Jilin University\\

    \{tszhang23, luyb24, caime24, zhezhang23\}@mails.jlu.edu.cn,\{liupp, zhangzijian, zhouqz\}@jlu.edu.cn
}

\usepackage{bibentry}

\begin{document}

\maketitle

\begin{abstract}
Current low-light image enhancement (LLIE) methods face significant limitations in simultaneously improving brightness while preserving semantic consistency, fine details, and computational efficiency. With the emergence of state-space models, particularly Mamba, image restoration has achieved remarkable performance, yet existing visual Mamba approaches flatten 2D images into 1D token sequences using fixed scanning rules, critically limiting interactions between distant tokens with causal relationships and constraining their ability to capture meaningful long-range dependencies. To address these fundamental limitations, we propose BSMamba, a novel visual Mamba architecture comprising two specially designed components: Brightness Mamba and Semantic Mamba. The Brightness Mamba revolutionizes token interaction patterns by prioritizing connections between distant tokens with similar brightness levels, effectively addressing the challenge of brightness restoration in LLIE tasks through brightness-guided selective attention. Complementing this, the Semantic Mamba establishes priority interactions between tokens sharing similar semantic meanings, allowing the model to maintain contextual consistency by connecting semantically related regions across the image, thus preserving the hierarchical nature of image semantics during enhancement. By intelligently modeling tokens based on brightness and semantic similarity rather than arbitrary scanning patterns, BSMamba transcends the constraints of conventional token sequencing while adhering to the principles of causal modeling. Extensive experiments demonstrate that BSMamba achieves state-of-the-art performance in LLIE while preserving semantic consistency. Code is available at 
https://github.com/bywlzts/BSMamba. 
\end{abstract}

\section{Introduction}
\label{sec:intro}

\begin{figure}[t]
    \centering
    \includegraphics[width=1\linewidth]{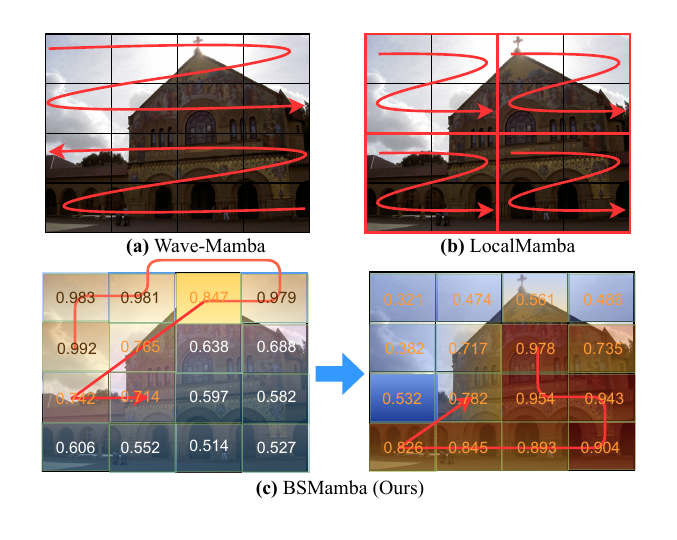}
    \caption{SS2D Scanning Strategies in State-Space Models. (a) The scanning structure of most visual Mamba models, such as Wave-Mamba, employs fixed horizontal, vertical, and reverse four-pass scanning strategies, which weaken interactions between causally related tokens. (b) The block-based scanning strategy, as in LocalMamba, also requires four scans within each block, lacking flexibility in interactions between causally related tokens. (c) Our BSMamba adopts a dual-layer scanning strategy based on brightness hierarchy scores and semantic hierarchy scores, allowing flexible interactions between causally related tokens while reducing scanning complexity.}
    \vspace{-0.3cm}
    \label{fig:mamba}
    \vspace{-0.1cm}
\end{figure}

Images captured in low-light conditions often suffer from low contrast, color distortion, detail loss, and complete information loss. Low Light Image Enhancement (LLIE) aims to restore brightness and visual content in such images, significantly expanding its applications in autonomous driving, security surveillance, and multimedia~\cite{hashmi2023featenhancer,du2024boosting,li2024light}.

Traditional LLIE methods, such as histogram equalization~\cite{lee2013contrast}, gamma correction~\cite{huang2012efficient}, and Retinex-based approaches~\cite{li2018structure, lol}, struggle in extreme lighting conditions due to generalization issues, often neglecting noise and detail degradation, which leads to unsatisfactory performance. In contrast, learning-based LLIE methods~\cite{retinexformer,lowlight8,zou2024vqcnir} learn the mapping between low-light and normal-light images through neural networks, offering superior performance and visual quality. However, many deep learning LLIE methods~\cite{four1,four2,zhang2024dmfourllie,xu2023low} overlook brightness consistency, color fidelity, and fine details during enhancement, resulting in color distortion, detail loss, and suboptimal brightness. Furthermore, due to parameter constraints, numerous LLIE methods~\cite{UHDFourICLR2023,wu2023learning} struggle to balance high performance with computational efficiency.

Recently, state-space models~\cite{gu2023mamba} have developed rapidly as alternatives to CNNs and Transformers, offering significant advantages in long sequence modeling while reducing computational complexity from quadratic to linear. Visual Mamba-based methods~\cite{huang2024irsrmamba,guo2025mambair,weng2025mamballie} have demonstrated superior performance over dominant Transformer models across various vision tasks, including image segmentation, super-resolution, and LLIE. However, applying Mamba to the visual domain still faces two critical limitations:

\textbf{First}, while causal relationships between tokens in language align well with state-space model scanning mechanisms, tokens with causal relationships in 2D images may be spatially distant, which weakens the modeling capabilities between these tokens and limits long-range modeling and generalization~\cite{xie2025quadmamba}. Additionally, flattening images into sequences disrupts the spatial dependencies and local adjacencies intrinsic to 2D images. This fundamental difference between language and vision domains poses challenges for adapting Mamba to visual tasks.

\textbf{Second}, current Mamba-based LLIE methods~\cite{zou2024wave,bai2024retinexmamba,weng2025mamballie} generally flatten 2D images into 1D sequences using fixed scanning rules, as illustrated in Fig.~\ref{fig:mamba}(a). To compensate for interactions between adjacent and distant tokens, these methods often employ multi-directional forward and reverse scanning strategies. However, MambaIR\cite{guo2025mambair,guo2024mambairv2} has revealed that this approach introduces significant information redundancy and fails to effectively address the decay of interactions among distant tokens. Various visual Mamba techniques~\cite{huang2024localmamba,ren2024autoregressive} have attempted to tackle this issue by segmenting images into multiple windows or clusters (Fig.~\ref{fig:mamba}(b)), processing each separately to group causally related tokens. Despite these efforts, these methods still lack flexibility in managing diverse object scales, leaving the challenge of long-range decay between causally related tokens unresolved.

To address these limitations, we introduce BSMamba, a novel Mamba architecture for LLIE tasks. BSMamba comprises Brightness Mamba and Semantic Mamba components. Brightness Mamba models the brightness essence of LLIE tasks by treating brightness levels as causal attributes between pixels, encouraging preferential interactions between spatially distant pixels with similar brightness properties, thereby enhancing long-range brightness-based causal interactions.  Semantic Mamba leverages instance segmentation results to generate attention maps. Unlike conventional attention mechanisms that treat all regions equally, Semantic Mamba dynamically assigns attention scores based on segmentation confidence scores while maintaining the relative importance of features within each instance. Semantic Mamba unfolds the image into a 1D sequence according to each object's attention score, encouraging priority interactions between distant tokens with similar semantic scores on a semantically consistent basis. Brightness Mamba and Semantic Mamba in BSMamba require only two scanning passes, improving efficiency while effectively enhancing brightness and semantic interactions between distant tokens, as shown in Fig.~\ref{fig:mamba}(c). After processing by BSMamba, the predicted image is finally fed into a Detail Enhancement Network (DE-Net) to further optimize image edge details. 

Overall, our contributions are summarized as follows: 
\begin{itemize}
\item We propose the BSMamba framework, which simultaneously focuses on brightness restoration, semantic consistency, and detail enhancement in LLIE tasks.
\item Brightness Mamba models long-range brightness causal interactions by prioritizing connections between spatially distant pixels with similar brightness levels, effectively enhancing brightness restoration.
\item Semantic Mamba establishes long-range semantic causal interactions by dynamically assigning attention scores based on instance segmentation, ensuring semantically consistent enhancement across distant regions.
\item BSMamba achieves state-of-the-art (SOTA) performance in LLIE tasks with improved efficiency and superior restoration of brightness, semantics, and details.
\end{itemize}

\begin{figure*}
    \centering
    \includegraphics[width=1\linewidth]{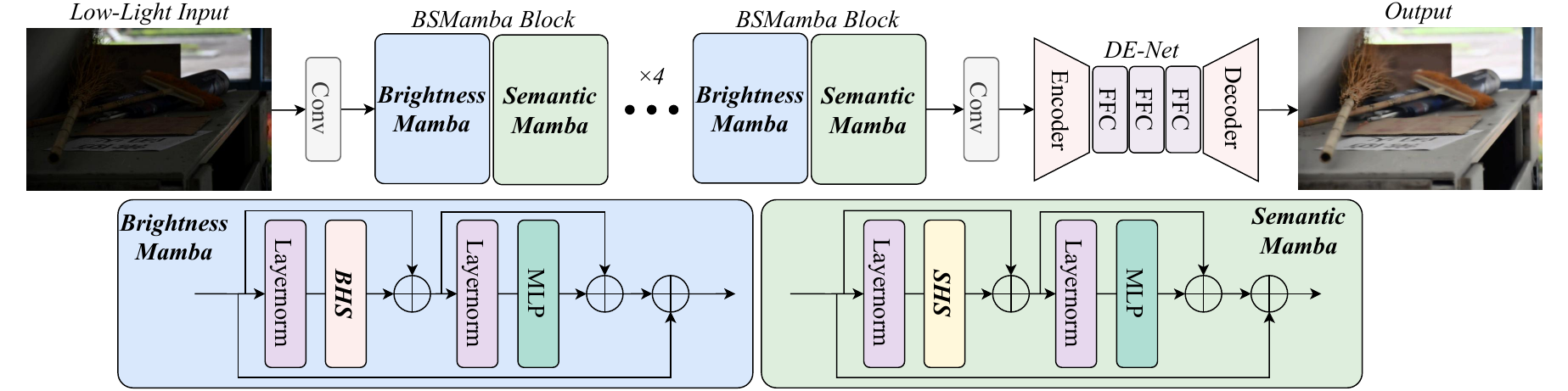}
    \vspace{-0.4cm}
    \caption{Overall Structure of BSMamba. BSMamba consists of four BSMamba blocks and the DE-Net.}
    \vspace{-0.2cm}
    \label{fig:BSMamba}
\end{figure*}

\section{Related Work}
\label{sec:relatedwork}

\subsection{Low-Light Scene Enhancement}
Traditional LLIE methods, such as histogram equalization, gamma correction, and Retinex theory, often introduce artifacts and noise under extreme lighting conditions. With the advent of deep learning, neural network-based LLIE methods~\cite{lowlight9,zhang2025cwnet,yan2025hvi} have become prevalent. Retinex-based approaches~\cite{wu2022uretinex}, including Retinexformer~\cite{retinexformer}, combine Retinex theory with Transformers to better manage long-range dependencies, while IAGC~\cite{wang2023low} introduces learnable gamma correction for adaptive enhancement.
Many methods also integrate priors to guide LLIE. For instance, SMG-LLE~\cite{xu2023low} uses structural maps, SKF~\cite{wu2023learning} incorporates semantic priors, and LEDN~\cite{wang2024multimodal} adds depth information to enhance restoration. However, these approaches face challenges related to errors from pre-trained models and increased computational costs due to the prior generation.
Recently, frequency-domain LLIE methods have emerged, such as FourLLIE~\cite{four1}, which enhances brightness through amplitude mapping in Fourier space, and DMFourLLIE~\cite{zhang2024dmfourllie}, which improves both phase and amplitude using infrared modalities and brightness attention. In addition, wave-based methods~\cite{zou2024wave,tan2024wavelet} apply state-space models in the wavelet domain for brightness enhancement and noise removal. However, these methods rely on standard Mamba, which fails to effectively model long-range interactions in 2D images.

\subsection{State-Space Models}
State-space models (SSMs) have gained traction for addressing long-range dependencies, as shown by LSSL~\cite{gu2021combining} and S4~\cite{gu2021efficiently}, which serve as alternatives to CNNs and Transformers. Mamba~\cite{gu2023mamba}, a selective SSM, has been adapted for visual tasks such as classification~\cite{zhu2024vision}, restoration~\cite{guo2025mambair,dong2024mamba}, and fusion~\cite{li2024mambadfuse}.
In LLIE, RetinexMamba~\cite{bai2024retinexmamba} merges Retinexformer with SSMs for improved processing speed, while WaveMamba~\cite{zou2024wave} and WalMaFa~\cite{tan2024wavelet} utilize Mamba for low-frequency feature extraction and global brightness modeling. Yet, these methods still depend on standard Mamba, which struggles with long-range interactions in 2D images.
To improve performance, several enhanced methods have been proposed. LocalMamba~\cite{huang2024localmamba} uses window-based scanning for better local detail, QuadMamba~\cite{xie2025quadmamba} employs quadtree-based scanning to maintain spatial locality, and ARM clusters adjacent blocks for efficient processing. LLEMamba~\cite{zhang2024llemamba} adopts bidirectional scanning to strike a balance between local and global focus. However, these approaches still struggle to model causal relationships between pixels, limiting effective interactions and overall performance.

\section{Method}
\label{sec:method}

\subsection{Overview}
The proposed BSMamba framework is designed to address the challenges of LLIE by simultaneously modeling brightness and semantic consistency. As illustrated in Fig.~\ref{fig:BSMamba}, the input low-light image is first processed through an initial convolutional layer to extract features. These feature maps are then refined by a series of BSMamba blocks, where the Brightness Mamba and Semantic Mamba modules enhance the interactions between causally related tokens based on brightness and semantic hierarchies. The refined feature maps are subsequently passed through a Detail Enhancement Network (DE-Net) to reconstruct spatial details, producing the final enhanced image.
\begin{figure}[t]
    \centering
    \includegraphics[width=1\linewidth]{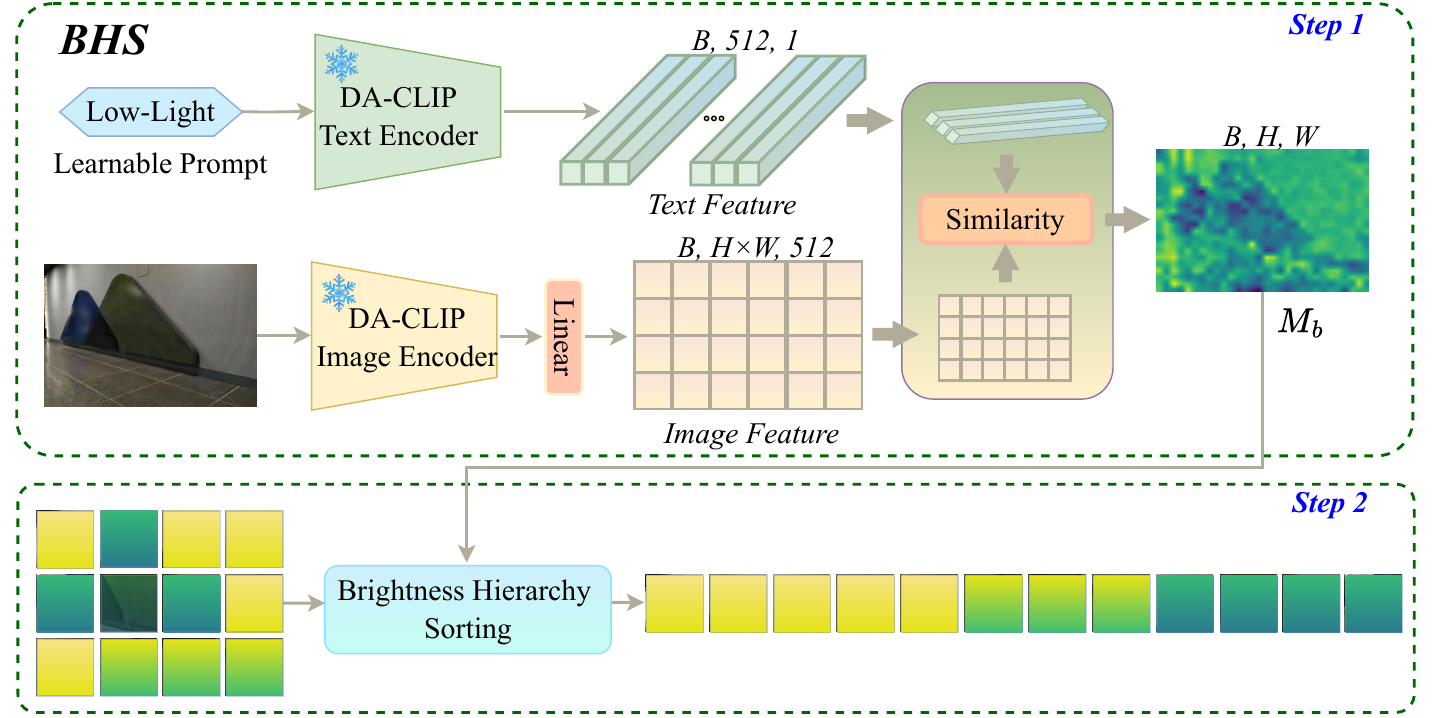}
    \vspace{-0.2cm}
    \caption{Brightness Hierarchy Scanning (BHS) in Brightness Mamba. The BHS structure enables brightness-aware token interactions by prioritizing tokens with similar brightness levels, regardless of their spatial positions.}
    \vspace{-0.2cm}
    \label{fig:bmamba}
\end{figure}

\subsection{Brightness Mamba}
To overcome the fundamental limitations of existing visual Mamba methods in modeling long-range dependencies, we propose Brightness Mamba, a novel component that prioritizes interactions between distant tokens with similar brightness levels. Specifically, Brightness Mamba utilizes DA-CLIP~\cite{luo2023controlling} to compute brightness grading scores by calculating similarity scores between all image pixels and brightness descriptors, reflecting the brightness levels of all pixels across the image. We unfold the image into a 1D sequence based on brightness scores, where pixels with similar brightness levels are grouped together, promoting effective interaction between spatially distant pixels with similar brightness characteristics.

This approach specifically targets the core requirements of LLIE tasks by enabling more effective brightness-guided enhancement across the image. As shown in Fig.~\ref{fig:BSMamba}, the proposed Brightness Mamba module integrates the Brightness Hierarchy Scanning (BHS) mechanism into the state-space structure:
\begin{equation}
\mathbf{X}' = \mathbf{X} +  \text{BHS}(\text{LayerNorm}(\mathbf{X})),
\end{equation}
\begin{equation}
\mathbf{Y} = \mathbf{X}' +  \text{MLP}(\text{LayerNorm}(\mathbf{X}')),
\end{equation}
where $\mathbf{X}$ is the input image. The BHS mechanism, as shown in Fig.~\ref{fig:bmamba}, consists of two key steps: 1) brightness hierarchy computation, and 2) brightness hierarchy sorting.  

\begin{figure}[t]
    \centering
    \includegraphics[width=1\linewidth]{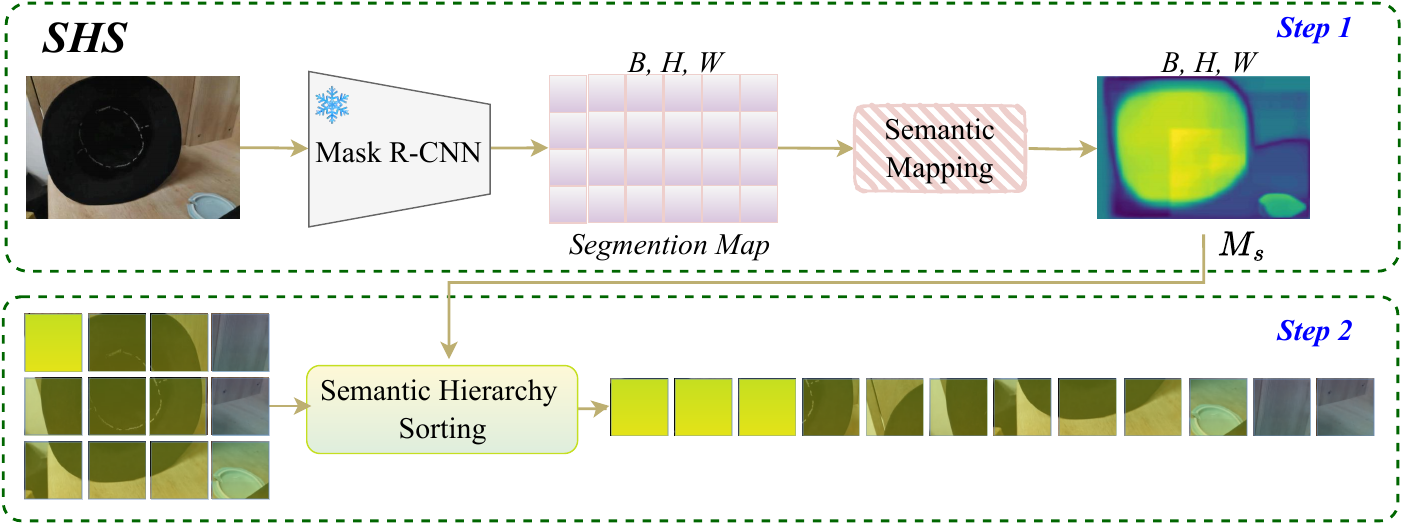}
    \vspace{-0.2cm}
    \caption{Semantic Hierarchy Scanning (SHS) in Semantic Mamba. SHS facilitates semantic-aware token interactions by grouping tokens based on their semantic significance, ensuring consistent enhancement within object instances.}
    \vspace{-0.2cm}
    \label{fig:smamba}
\end{figure}

\textbf{Step 1: Brightness Hierarchy Computation.}  
Traditional visual Mamba methods scan images using fixed patterns, which fail to capture meaningful causal relationships between distant pixels. In LLIE tasks, pixels with similar brightness properties often require similar enhancement operations, regardless of their spatial distance. DA-CLIP predicts high-quality feature embeddings from degraded low-light images. Therefore, we leverage a pre-trained DA-CLIP model to compute brightness similarity scores, enabling brightness-aware token interactions:  
\vspace{-0.1cm}
\begin{equation}
\mathbf{M}_{b} = \text{Sim}(\Phi_{image}(\mathbf{X}), \Phi_{text}(\text{`low-light'})),
\end{equation}
where $\Phi_{\text{image}}$ extracts image features from the DA-CLIP image encoder, followed by projection to align feature dimensions and normalization for cosine similarity computation. $\Phi_{\text{text}}$ encodes the text descriptor `low-light.' The resulting similarity map quantifies the alignment of each pixel with the concept of `low-light,' effectively creating a brightness hierarchy map $\mathbf{M}_{b} \in \mathbb{R}^{B \times H \times W}$ across the image.

\textbf{Step 2: Brightness Hierarchy Sorting.}  
Based on the brightness hierarchy map $\mathbf{M}_{b}$ generated in Step 1, we perform Brightness Hierarchy Sorting to reorder the image tokens according to their brightness levels:  

\begin{equation}
\hat{\mathbf{X}}_{\text{sort}}, \hat{\mathbf{I}}_{\text{sort}} = \text{Sort}(\mathbf{M}_{b}, \mathbf{X}),
\end{equation}

where $\hat{\mathbf{X}}_{\text{sort}}$ represents the 1D sequence of tokens reordered based on their brightness scores, ensuring that pixels with similar brightness characteristics are processed sequentially, regardless of their spatial positions in the original image. $\hat{\mathbf{I}}_{\text{sort}}$ is the inverse index used to reorder the tokens back to their original positions. Tokens from different spatial locations but with similar brightness levels (indicated by blocks of similar colors) are grouped together in the sorted sequence. By prioritizing interactions between tokens with similar brightness levels, Brightness Mamba effectively models the causal relationships of brightness across the image space, addressing the core challenges of LLIE.

\subsection{Semantic Mamba}

While addressing brightness consistency is crucial for LLIE tasks, maintaining semantic coherence between object instances is equally essential to ensure perceptually natural results. To accomplish this, we propose the Semantic Mamba module, as shown in Fig.\ref{fig:BSMamba}, which leverages instance segmentation information to guide token interactions based on semantic relationships.  
Similar to Brightness Mamba, Semantic Mamba incorporates the SHS mechanism within a state-space block structure.  
As shown in Fig.~\ref{fig:smamba}, the Semantic Hierarchy Scanning (SHS) mechanism consists of two key steps: 1) semantic mapping using Mask R-CNN~\cite{he2017mask}, and 2) semantic hierarchy sorting.  

\textbf{Step 1: Semantic Mapping.}   
To effectively capture semantic information from low-light images, we utilize a pre-trained Mask R-CNN model to generate instance segmentation masks. Unlike traditional scanning patterns that fail to respect semantic boundaries, our approach identifies object instances and assigns attention scores based on their semantic significance:  
\begin{equation}  
\mathbf{M}_s = \mathcal{M}_{\text{semantic}}({G}_{\text{Mask-RCNN}}(\mathbf{X})) , 
\end{equation}  
where $\mathbf{X}$ is the input image, ${G}_{\text{Mask-RCNN}}$ generates instance segmentation masks with confidence scores, and $\mathcal{M}_{\text{semantic}}$ maps these instances to a continuous semantic attention map $\mathbf{M}_s \in \mathbb{R}^{B \times H \times W}$.  
The semantic mapping function $\mathcal{M}_{\text{semantic}}$ implements a dynamic grading strategy:  
\vspace{-0.1cm}
\begin{equation}
\mathcal{M}_{\text{semantic}}(\mathbf{I}) = \sum_{i=1}^{n} \mathbf{M}_i \cdot \mathbf{S}_i \cdot \mathcal{G}_i + \mathbf{M}_{\text{bg}} \cdot \mathcal{G}_{\text{bg}},
\end{equation}
where $\mathbf{I}$ represents the segmentation results, $n$ is the number of detected instances, $\mathbf{M}_i$ and $\mathbf{S}_i$ are the mask and confidence score for the $i$-th instance, $\mathcal{G}_i$ is the grading function that maps the $i$-th instance to a specific attention range, $\mathbf{M}_{\text{bg}}$ is the background mask, and $\mathcal{G}_{\text{bg}}$ maps the background to the lowest attention range.  Specifically, we define $n+1$ hierarchical ranges (including the background) as:  

\begin{equation}  
\mathcal{G}_i = \left[\frac{i}{n+1}, \frac{i+1}{n+1}\right], i \in \{0, 1, \ldots, n\} , 
\end{equation}  
where $i=0$ corresponds to the background. This dynamic grading ensures that:  
1) Each detected instance is assigned a unique attention range, with higher confidence instances receiving higher attention values.  
2) The background regions receive the lowest attention values.  
3) Different parts within the same instance maintain their relative importance.

\begin{table}[t]  
\centering  
\renewcommand{\arraystretch}{1.2} 
\setlength{\tabcolsep}{4pt}
\resizebox{1.0\columnwidth}{!}{
\begin{tabular}{l|c|ccc|c}  
\toprule  
\textbf{Methods} & \textbf{Venue} & \multicolumn{3}{c|}{\textbf{LSRW-Nikon}} & \textbf{\#Param} \\   
                 &                 & \textbf{PSNR ↑} & \textbf{SSIM ↑} & \textbf{LPIPS ↓} & \textbf{(M)} \\ \midrule  
MIRNet           & ECCV, 20       & 17.10           & 0.5022          & 0.2170          & 31.79         \\
Kind++           & IJCV, 21            & 14.79           & 0.4749          & \underline{0.2111}          & 8.27          \\
SGM              & TIP, 21          & 15.73           & 0.4971          & 0.2234          & 2.31          \\
FECNet           & ECCV, 22       & 17.06           & 0.4999          & 0.2192          & 0.15         \\
HDMNet           & MM, 22          & 16.65           & 0.4870          & 0.2157          & 2.32          \\
FourLLIE        & MM, 23             & 17.82           & 0.5036          & 0.2150          & 0.12         \\
UHDFour         & ICLR, 23           & \underline{17.94} & 0.5195 & 0.2546          & 17.54         \\
Retinexformer    & ICCV, 23           & 17.64           & 0.5082          & 0.2784          & 1.61          \\ 
WaveMamba       & MM, 24          & 17.34           & 0.5192          & 0.2294          & 1.26          \\ UHDFormer    & AAAI, 24       & 17.01           & 0.5186         & 0.2793          & 0.34 \\
DMFourLLIE      & MM, 24            & 17.04           & \underline{0.5294} & 0.2178          & 0.75          \\
RetinexMamba   & ICONIP, 24       & 17.59          & 0.5133          & 0.2534          & 3.59        \\
MambaLLIE       & NIPS, 24             & 17.25           & 0.5084          & 0.2612          & 2.28          \\
CWNet           & ICCV, 25          & 17.38           & 0.5119          & 0.2732          & 1.23          \\
CIDNet          & CVPR, 25              & 17.16           & 0.4975          & 0.2859          & 1.88          \\ \midrule
Ours             &  & \textbf{18.04} & \textbf{0.5327} & \textbf{0.2012}  & 0.68  \\ 
\bottomrule  
\end{tabular}}  
\vspace{-0.2cm}
\caption{Quantitative comparison on LSRW-Nikon dataset. The best and second-best results are highlighted in bold and underlined, respectively.}  
\vspace{-0.2cm}
\label{tab:com-nikon}  
\end{table}

\begin{figure*}[t]
    \centering
    \includegraphics[width=1\linewidth]{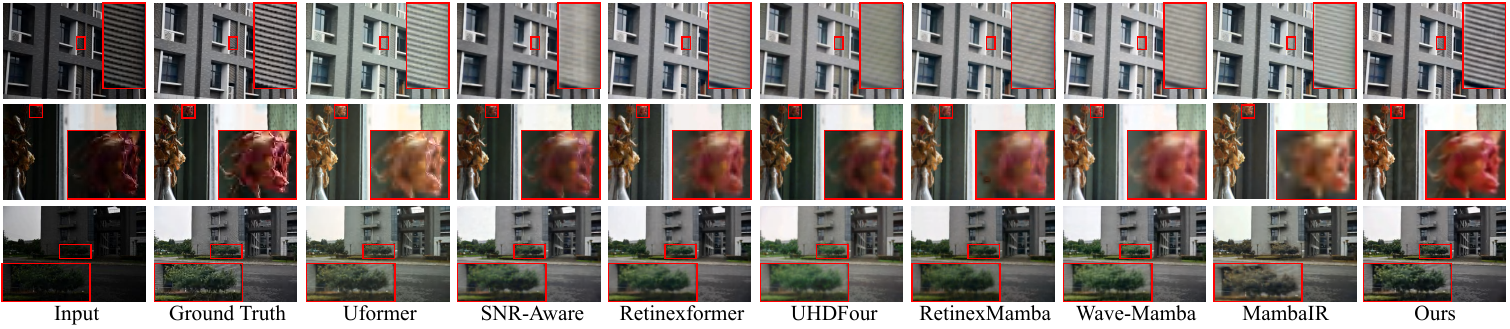}
    \vspace{-0.4cm}
    \caption{Visual comparison on LSRW-Nikon dataset. Zoomed-in provide a clearer comparison.}
    \vspace{-0.2cm}
    \label{fig:nikon}
\end{figure*}

\begin{figure*}[t]
    \centering
    \includegraphics[width=1\linewidth]{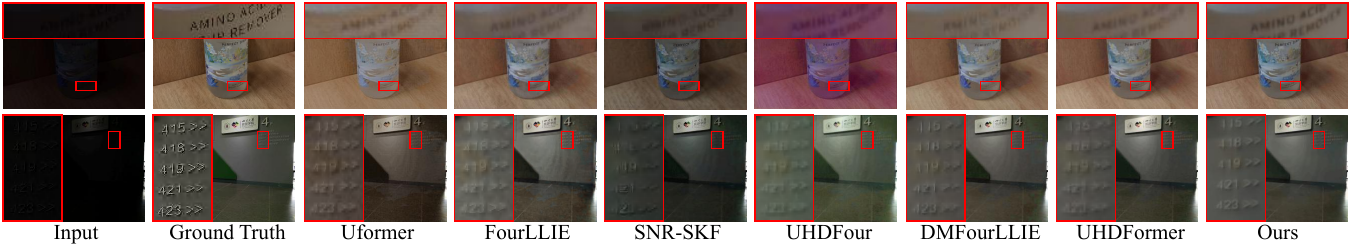}
    \vspace{-0.4cm}
    \caption{Visual comparison on LSRW-Huawei dataset. Zoomed-in provide a clearer comparison.}
    \vspace{-0.2cm}
    \label{fig:huawei}
\end{figure*}

\begin{figure*}[t]
    \centering
    \includegraphics[width=1\linewidth]{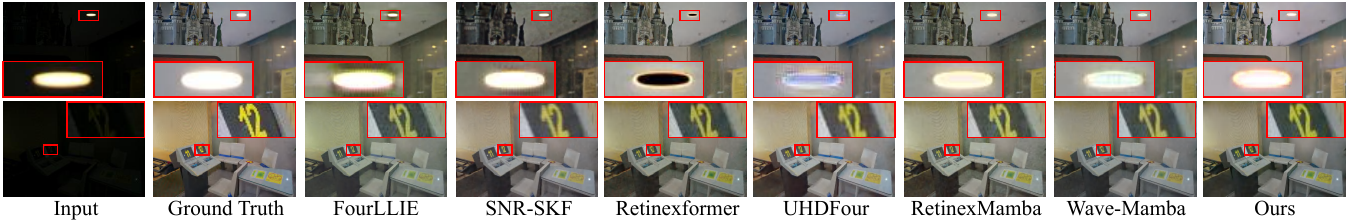}
    \vspace{-0.4cm}
    \caption{Visual comparison on LOL dataset. Zoomed-in provide a clearer comparison.}
    \vspace{-0.2cm}
    \label{fig:lol}
\end{figure*}

\begin{table}[t]  
\centering  
\renewcommand{\arraystretch}{1.2} 
\setlength{\tabcolsep}{4pt} 
\resizebox{1.0\columnwidth}{!}{ 
\begin{tabular}{l|c|ccc|c}  
\toprule  
\textbf{Methods} & \textbf{Venue} & \multicolumn{3}{c|}{\textbf{LSRW-Huawei}} & \textbf{\#FLOPs} \\   
                 &                 & \textbf{PSNR ↑} & \textbf{SSIM ↑} & \textbf{LPIPS ↓} & \textbf{(G)} \\ \midrule  
MIRNet        & ECCV, 20       & 19.98           & 0.6085          & 0.2154          & 785.1        \\
Kind++         & IJCV, 21        & 15.43           & 0.5695          & 0.2366          & -            \\
SGM           & TIP, 21        & 18.85           & 0.5991          & 0.2492          & -            \\
FECNet       & ECCV, 22       & 21.09           & 0.6119          & 0.2341          & 5.82         \\
HDMNet         & MM, 22         & 20.81           & 0.6071          & 0.2375          & -            \\
SNR-Aware     & ICIP, 22       & 20.67           & 0.5911          & 0.1923          & 26.35        \\
FourLLIE     & MM, 22         & 21.11           & 0.6256          & 0.1825          & 4.07         \\
UHDFour       & ICLR, 23       & 19.39           & 0.6006          & 0.2466          & 4.78         \\
Retinexformer   & ICCV, 23       & 21.23           & 0.6309          & 0.1699          & 15.57        \\ 
WaveMamba    & MM, 24         & 21.19           & 0.6391          & 0.1818          & 7.22         \\ 
UHDFormer    & AAAI, 24       & 20.64           & 0.6244          & 0.1812          & 3.24         \\ 
DMFourLLIE      & MM, 24         & 21.47           & 0.6331          & 0.1781          & 5.81         \\ 
RetinexMamba   & ICONIP, 24       & 20.88           & 0.6298          & 0.1689          & 34.76        \\
MambaLLIE      & NIPS, 24       & 20.98           & 0.6388          & 0.1894          & 20.85        \\
CWNet      & ICCV, 25 & \textbf{21.50} & \underline{0.6397} & \underline{0.1682} & 11.3         \\
CIDNet      & CVPR, 25        & 20.30           & 0.6054          & 0.1973          & 7.57         \\ \midrule
Ours             &       & \underline{21.49} & \textbf{0.6435} & \textbf{0.1613} & 8.15 \\ 
\bottomrule  
\end{tabular}}  
\vspace{-0.2cm}
\caption{Quantitative comparison on LSRW-Huawei dataset.}  
\vspace{-0.2cm}
\label{tab:com-huawei}  
\end{table}

\begin{table}[t]  
\centering  
\renewcommand{\arraystretch}{1.2} 
\setlength{\tabcolsep}{4pt} 
\resizebox{\columnwidth}{!}{
\begin{tabular}{l|ccc|ccc}  
\toprule  
\textbf{Methods} & \multicolumn{3}{c|}{\textbf{LOL-v1}} & \multicolumn{3}{c}{\textbf{LOL-v2-Real}} \\   
                 & \textbf{PSNR ↑} & \textbf{SSIM ↑} & \textbf{LPIPS ↓} & \textbf{PSNR ↑} & \textbf{SSIM ↑} & \textbf{LPIPS ↓} \\ \midrule  
FourLLIE          & 20.99 & 0.8071  & 0.0952 & 23.45 & 0.8450  & 0.0613    \\              
UHDFour         & 22.89 & 0.8147  & 0.0934 & 27.27 & 0.8579  & 0.0617 \\
SNR-SKF        & 20.51 & 0.7110  & 0.1810 & 21.82 & 0.7468  & 0.1513 \\
Retinexformer         & 22.71 & 0.8177  & 0.0922 & 24.55 & 0.8434  & 0.0627 \\
RetinexMamba        & 23.15 & 0.8210  & 0.0876 & 27.31 & 0.8667  & 0.0551 \\
WaveMamba       & 22.76 & 0.8419  & 0.0791 & \underline{27.87} & 0.8935  & 0.0451 \\
DMFourLLIE        & 21.58 & 0.8349  & 0.0821 & 26.74 & 0.8792  & 0.0528 \\
MambaLLIE     & 22.80 & 0.8315& 0.8450 & 26.96 & 0.8875 & 0.0483 \\ 
CIDNet  & 23.81 & \textbf{0.8574} & 0.0856 & - & - & - \\ 
CWNet  & 23.60 & 0.8496 & \textbf{0.0648} &27.39 & \underline{0.9005} & \textbf{0.0383} \\
 \midrule   
Ours            & \textbf{24.07}  & \underline{0.8521} & \underline{0.0709} & \textbf{29.07}  & \textbf{0.9013} & \underline{0.0401} \\ \bottomrule  
\end{tabular}}  
\vspace{-0.2cm}
\caption{Quantitative comparison on LOL dataset.}  
\label{tab:com-lol}  
\vspace{-0.2cm}
\end{table}

\textbf{Step 2: Semantic Hierarchy Sorting.} 
Based on the semantic attention map $\mathbf{M}_s$ generated in Step 1, we perform Semantic Hierarchy Sorting to reorder image tokens according to their semantic significance:  
\begin{equation}  
\tilde{\mathbf{X}}_{\text{sort}}, \tilde{\mathbf{I}}_{\text{sort}}  = \text{Sort}(\mathbf{M}_s, \mathbf{X})  ,
\end{equation}  
similar to the BHS, where $\tilde{\mathbf{X}}_{\text{sort}}$ represents the tokens reordered based on their semantic scores, ensuring that tokens from the same semantic entity are processed together regardless of their spatial locations. $\tilde{\mathbf{I}}_{\text{sort}}$ contains the indices to reorder tokens back to their original positions.  As illustrated in Fig.~\ref{fig:smamba}, tokens with similar semantic significance (represented by similar colors) are grouped together in the sorted sequence. This approach ensures that:  
1) Pixels belonging to the same object instance are processed sequentially.    
2) Tokens with similar semantic roles can interact effectively regardless of their spatial distance.  

The Semantic Mamba complements the Brightness Mamba within the BSMamba framework. While Brightness Mamba focuses on enhancing the brightness consistency of the image based on illumination-related features, Semantic Mamba ensures that the semantic relationships between different image regions are preserved during enhancement. This dual-perspective approach addresses both critical aspects of LLIE:  
Brightness Mamba establishes long-range dependencies between pixels with similar illumination conditions.  
Semantic Mamba maintains semantic coherence by ensuring consistent enhancement within object instances.  

\subsection{Detail Enhancement Network (DE-Net)}
After processing with BSMamba blocks, which restore brightness relationships and semantic consistency, we further refine the spatial details and texture information through our proposed Detail Enhancement Network (DE-Net). As illustrated in Fig.~\ref{fig:BSMamba}, DE-Net is specifically designed to enhance fine-grained details that may remain partially degraded after the initial enhancement. The DE-Net adopts an encoder-decoder architecture with Fast Fourier Convolution (FFC)~\cite{chi2020fast} blocks in the bottleneck. The FFC architecture divides feature channels into spatial and spectral branches, allowing the network to simultaneously capture local spatial details and global frequency patterns, which is particularly beneficial for enhancing texture and structural details~\cite{dong2022incremental}.  

To further enhance the structural integrity of the reconstructed images, we incorporate edge guidance through a specialized Canny edge loss:  

\begin{equation}  
\mathcal{L}_{\text{canny}} = \text{BCE}(\mathcal{E}(\mathbf{Y}), \mathcal{E}(\mathbf{Y}_{\text{gt}})),  
\end{equation}  
where $\mathcal{E}$ represents the Canny edge extraction operation, $\mathbf{Y}$ is the predicted enhanced image, $\mathbf{Y}_{\text{gt}}$ is the ground truth image, and $\text{BCE}$ denotes the binary cross-entropy loss function~\cite{ruby2020binary}. 

The overall optimization objective is: 

\begin{equation}  
\mathcal{L}_{\text{total}} = \lambda_{1}\mathcal{L}_{\text{1}}  +\lambda_{2}\mathcal{L}_{\text{ssim}}   +\lambda_{3}\mathcal{L}_{\text{canny}}  ,
\end{equation}  
where loss weights $\lambda_{1}, \lambda_{2}, \lambda_{3} = [1.0, 0.5, 0.1]$, $\mathcal{L}_{\text{1}}$ is the $l_{1}$ loss, $\mathcal{L}_{\text{ssim}}$ denotes the SSIM loss~\cite{SSIM}.

\section{Experiments}


\begin{figure}[t]
    \centering
    \includegraphics[width=0.9\linewidth]{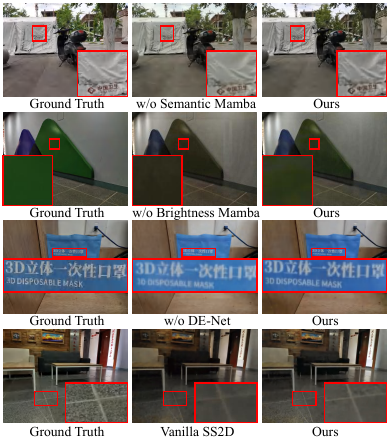}
    \vspace{-0.2cm}
    \caption{Visualization of Ablation Studies.}
    \vspace{-0.2cm}
    \label{fig:ab_show}
\end{figure}

\begin{figure}[t]
    \centering
    \includegraphics[width=0.9\linewidth]{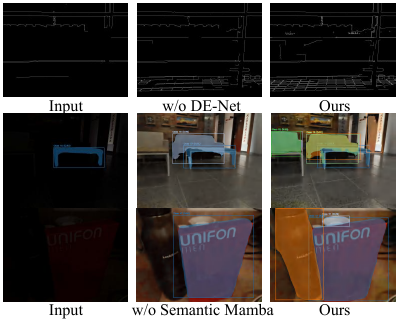}
    \vspace{-0.2cm}
    \caption{Validation on Downstream Applications. We validate DE-Net's effectiveness through edge detection, which is sensitive to image details, and Semantic Mamba's effectiveness through instance segmentation.}
    \vspace{-0.2cm}
    \label{fig:ab_seg}
\end{figure}

\begin{figure}[t]
    \centering
    \includegraphics[width=1.0\linewidth]{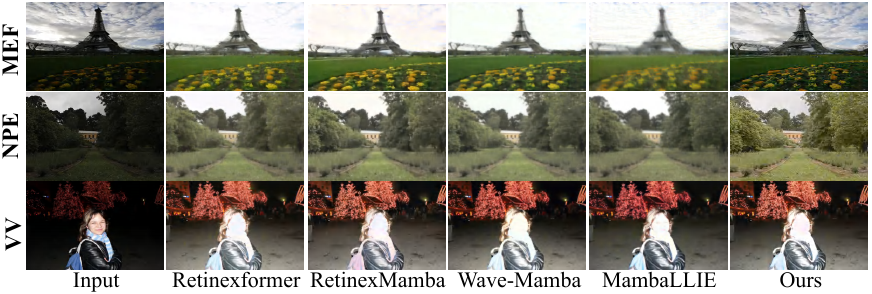}
    \vspace{-0.2cm}
    \caption{Visual comparisons on unpair datasets.}
    \vspace{-0.2cm}
    \label{fig:small}
\end{figure}

\subsection{Datasets and Experimental Setting}  
Our method is evaluated on four widely recognized datasets for LLIE: LOL~\cite{lol}, LOLv2-Real, LSRW-Huawei~\cite{lsrw}, and LSRW-Nikon, comprising a total of 485 training pairs and 15 test pairs for LOL, 3,150 training pairs and 20 test pairs for LSRW-Huawei, and 2,450 training pairs and 30 test pairs for LSRW-Nikon. Additionally, we assess performance on three unpaired datasets: NPE~\cite{npe}, MEF~\cite{mefl}, and VV.

The model is implemented and trained end-to-end using the PyTorch framework. During training, images are randomly cropped to $256 \times 256$ pixels and augmented with random flips. We optimize the model using the ADAM optimizer with an initial learning rate of $4.0\times10^{-4}$ and a multi-step learning rate scheduler. The batch size is set to $8$, and training spans $1.5\times10^{5}$ iterations, conducted on two NVIDIA 4090 GPUs.

\subsection{Overall Performance}  
We compare our approach against various state-of-the-art LLIE methods, including learning-based techniques (Kind~\cite{kind}, Kind++~\cite{kind++}, MIRNet~\cite{lowlight9}, SGM~\cite{lol}, HDMNet~\cite{liang2022learning}, SNR-Aware~\cite{lowlight8}, SNR-SKF~\cite{wu2023learning}, Retinexformer~\cite{retinexformer}, UHDFormer~\cite{wang2024uhdformer}, CIDNet~\cite{yan2025hvi}), Fourier-based models (FourLLIE~\cite{four1}, FECNet~\cite{four2}, UHDFour~\cite{UHDFourICLR2023}, DMFourLLIE~\cite{zhang2024dmfourllie}), and Mamba-based approaches (RetinexMamba~\cite{bai2024retinexmamba}, Wave-Mamba~\cite{zou2024wave}, MambaIR~\cite{guo2025mambair}, MambaLLIE~\cite{weng2025mamballie}, CWNet~\cite{zhang2025cwnet}). For fair evaluation, all deep learning models are trained on the same datasets using their publicly available implementations. We use Peak Signal-to-Noise Ratio (PSNR), Structural Similarity Index (SSIM)~\cite{SSIM}, and Learned Perceptual Image Patch Similarity (LPIPS)~\cite{LP} as evaluation metrics.

\begin{table}[t]  
\centering  
\renewcommand{\arraystretch}{1.2} 
\setlength{\tabcolsep}{4pt} 
\resizebox{1.0\columnwidth}{!}{ 
\begin{tabular}{l|ccc}  
\toprule  
\textbf{Configurations} & \textbf{PSNR ↑} & \textbf{SSIM ↑} & \textbf{LPIPS ↓} \\ \midrule  
w/o Brightness Mamba    & 21.03    & 0.6357    & 0.1893   \\  
w/o Semantic Mamba    & 21.14  & 0.6374 & 0.1795     \\ 
w/o DE-Net                & 20.94      & 0.6389   & 0.1747  \\
BSMamba Block $\rightarrow$ Vanilla SS2D      & 20.57  & 0.6351 & 0.1905  \\ \midrule 
Brightness Mamba \textbf{+} Semantic Mamba & 20.57   & 0.6351  & 0.1905\\
Brightness Mamba \textbf{C} Semantic Mamba& 20.57   & 0.6351    & 0.1905\\
Semantic Mamba $\leftrightarrow$ Brightness Mamba& 21.34   & 0.6421    & 0.1682 \\ \midrule
DA-CLIP $\rightarrow$ Brightness Histogram & 21.36   & 0.6402    & 0.1665\\
DA-CLIP $\rightarrow$ Y component& 21.23   & 0.6384    & 0.1714 \\ \midrule
$\lambda_{1}, \lambda_{2}, \lambda_{3} = [1.0, 0.3, 0.1]$ & \textbf{21.51}   & 0.6408    & \underline{0.1637}\\ 
$\lambda_{1}, \lambda_{2}, \lambda_{3} = [1.0, 0.5, 0.2]$ & 21.29   & \underline{0.6431}    & 0.1688\\ \midrule
Ours             & \underline{21.49} & \textbf{0.6435} & \textbf{0.1613}          \\
\bottomrule  
\end{tabular}}  
\vspace{-0.2cm}
\caption{Ablation studies and experimental configurations.}  
\vspace{-0.2cm}
\label{tab:ablation}  
\end{table}

\noindent \textbf{Quantitative Results on LOL, LOLv2-Real, LSRW-Huawei and LSRW-Nikon.}  
Our method demonstrates advantages in model size and efficiency. \textbf{1)} Tab.~\ref{tab:com-nikon} presents results on the LSRW-Nikon dataset, where our approach significantly outperforms existing methods. \textbf{2)} In Tab.~\ref{tab:com-huawei}, we achieved competitive results on the LSRW-Huawei dataset, with our PSNR ranking second, just 0.01 dB behind CWNet. \textbf{3)} Tab.~\ref{tab:com-lol} shows that we also perform competitively on the LOL-v1 and LOL-v2-Real datasets.

\noindent \textbf{Qualitative Comparison.}  
To illustrate restoration quality, we provide visual comparisons with state-of-the-art methods.
\textbf{1)} Fig.\ref{fig:nikon} shows results on the LSRW-Nikon dataset, where existing methods suffer from blurring and color distortion, especially in fine textures. In contrast, our approach maintains color fidelity and clear texture details, aligning closely with the Ground Truth.
\textbf{2)} Fig.\ref{fig:huawei} presents comparisons on the LSRW-Huawei dataset. Our method excels in restoring text and object textures in challenging regions, with a more natural color distribution.
\textbf{3)} Fig.~\ref{fig:lol} illustrates results on the LOL dataset. Other methods misinterpret bright light sources, leading to color shifts, while our approach preserves true brightness and effectively enhances dark areas, ensuring a natural luminance balance and reducing artifacts.


\subsection{Ablation Study and Experimental Configuration}  
\noindent \textbf{Visual Ablation:}  
As shown in Fig.~\ref{fig:ab_show}, removing Brightness Mamba significantly compromises brightness enhancement. Excluding Semantic Mamba reduces semantic consistency and instance-level performance, while removing DE-Net leads to blurred details and decreased edge sharpness. Replacing our Brightness and Semantic Mamba scanning strategies with Vanilla SS2D (fixed multi-directional scanning) in the BSMamba Block results in severe declines in both brightness and image details.

\noindent \textbf{Quantitative Ablation:}  
Tab.~\ref{tab:ablation} presents our ablation studies. The first set modifies and replaces core modules; the second set tests combinations of the two Mamba modules (addition, concatenation, reordering) to validate our optimal strategy. The third set assesses the impact of changing DA-CLIP’s brightness score to histogram and YCrCb luminosity statistics. The fourth group examines the effects of varying loss weights. These four groups comprehensively validate the effectiveness of our components and configurations.

\noindent \textbf{Downstream Applications:}
Beyond the brightness-focused Brightness Mamba, we introduce Semantic Mamba and DE-Net to enhance semantic consistency and detail restoration. To validate their effectiveness, we evaluate our method on downstream tasks, including instance segmentation (Mask R-CNN~\cite{he2017mask} pre-trained on COCO dataset~\cite{lin2014microsoft}) and edge detection~\cite{ding2001canny}, which assess semantic preservation and detail recovery, respectively. Fig.\ref{fig:ab_show} illustrates the results of removing the corresponding modules and compares them. The visual comparisons further confirm the performance gains achieved by incorporating Semantic Mamba and DE-Net.

\subsection{Limitations and Conclusion}  
\noindent \textbf{Limitations:}  
As observed in Fig.~\ref{fig:small}, our visual comparisons with current SOTA Mamba-based methods on paired datasets show that while we achieve favorable color and detail in the first two rows, we still encounter exposure issues in the third row. This insight motivates us to address the challenge of suppressing exposure artifacts while ensuring color semantic and brightness recovery in future work.

\noindent \textbf{Conclusion:}  
In this paper, we introduced BSMamba, a novel visual Mamba architecture aimed at overcoming the fundamental limitations of LLIE. By incorporating brightness and semantic Mamba, our approach redefines the label interaction paradigm based on brightness and semantic similarity, enabling effective long-range modeling of causally related labels. In the future, we plan to explore generalized scanning strategies or develop customized scanning techniques tailored for specific downstream tasks to enhance performance.
\bibliography{aaai2026}



\end{document}